\documentclass[10pt,twocolumn,letterpaper]{article}
\usepackage[accsupp]{axessibility}
\usepackage{iccv}
\usepackage{times}
\usepackage{epsfig}
\usepackage{graphicx}
\usepackage{amsmath}
\usepackage{amssymb}
\usepackage{multirow}
\usepackage{bbm}
\usepackage{color}
\usepackage[ruled]{algorithm2e}

\newcommand*\samethanks[1][\value{footnote}]{\footnotemark[#1]}

\usepackage[pagebackref=true,breaklinks=true,letterpaper=true,colorlinks,bookmarks=false]{hyperref}

\iccvfinalcopy 

\def\iccvPaperID{5652} 

\ificcvfinal\fi

\begin{document}

\title{Vision-Language Navigation with Random Environmental Mixup}
 
\author{Chong Liu \textsuperscript{1,2}\thanks{Equal Contribution}  \; Fengda Zhu \textsuperscript{3}\samethanks
\; Xiaojun Chang \textsuperscript{4} \; Xiaodan Liang\textsuperscript{5} \; Zongyuan Ge\textsuperscript{3} \; Yi-Dong Shen \textsuperscript{1}\thanks{Corresponding author} \\
\textsuperscript{1} State Key Laboratory of Computer Science, Institute of Software, Chinese Academy of Sciences, China \\
\textsuperscript{2} University of Chinese Academy of Sciences, Beijing 100049, China \\
\textsuperscript{3} Monash University, Melbourne, Australia\\
\textsuperscript{4} RMIT University, Melbourne, Australia\\
\textsuperscript{5} Sun Yat-sen University, Guangzhou, China\\
{\tt\small liuchong@ios.ac.cn \ fengda.zhu@monash.edu \ xiaojun.chang@rmit.edu.au}\\
{\tt\small xdliang328@gmail.com \ zongyuan.ge@monash.edu \  ydshen@ios.ac.cn}
}

\maketitle

\ificcvfinal\thispagestyle{empty}\fi

\begin{abstract}
Vision-language Navigation (VLN) tasks require an agent to navigate step-by-step while perceiving the visual observations and comprehending a natural language instruction.
Large data bias, which is caused by the disparity ratio between the small data scale and large navigation space, makes the VLN task challenging.
Previous works have proposed various data augmentation methods to reduce data bias. However, these works do not explicitly reduce the data bias across different house scenes. 
Therefore, the agent would overfit to the seen scenes and 
achieve poor navigation performance in the unseen scenes.
To tackle this problem, we propose the Random Environmental Mixup (REM) method, which generates cross-connected house scenes as augmented data via mixuping environment. Specifically, we first select key viewpoints according to the room connection graph for each scene. Then, we cross-connect the key views of different scenes to construct augmented scenes. Finally, we generate augmented instruction-path pairs in the cross-connected scenes.
The experimental results on benchmark datasets demonstrate that our augmentation data via REM help the agent reduce its performance gap between the seen and unseen environment and improve the overall performance, making our model the best existing approach on the standard VLN benchmark. The code have released: https://github.com/LCFractal/VLNREM.
\end{abstract}

\section{Introduction}

Recently, there is a surge of research interests in Vision-Language Navigation (VLN)~\cite{Seq2Seq} tasks, in which an agent learns to navigate by following a natural language instruction. 
The agent begins at a random point and goes towards a goal via actively exploring the environments.
Before the navigation starts, the agent receives a language instruction. 
At every step, the agent can get the surrounding visual information.
The key to this task is to perceive the visual scene and comprehend natural language instructions sequentially and make actions step-by-step. 

\begin{figure}[t]
   \begin{center}
   \includegraphics[width=0.9\linewidth]{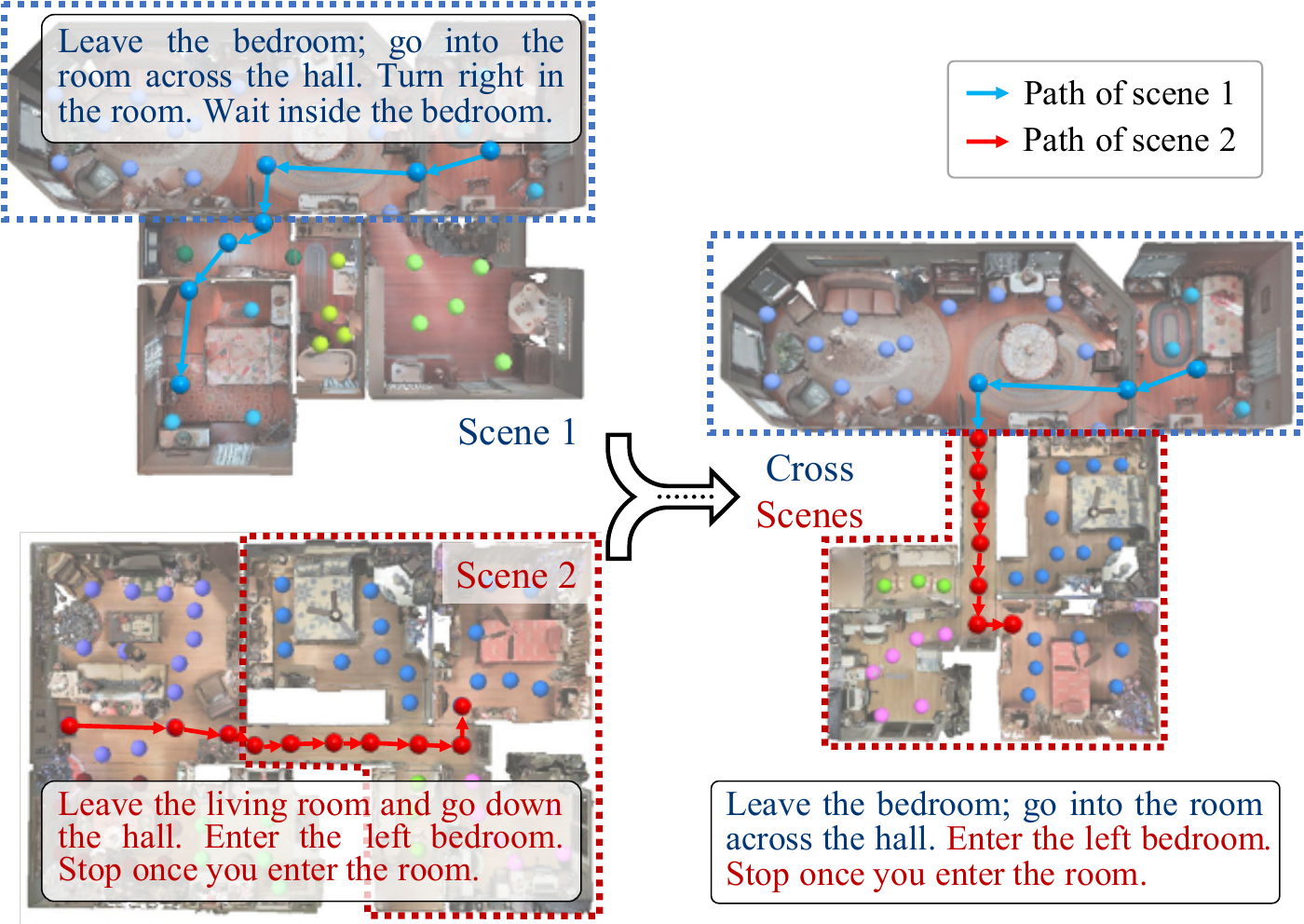}
   \end{center}
  \vspace{-1 em}
      \caption{\textbf{REM} mixes up two scenes and generates data triplets (environment, path, instruction). We divide the two scenes and recombine them to construct a new cross-connected scene, and reconstruct the corresponding paths and instructions.} 
   \label{fig:introduction}
   \vspace{-1 em}
\end{figure}

Recent advances made by deep learning works in the domains of feature extraction~\cite{he2016deep, anderson2018bottom, mikolov2013distributed, ren2015faster}, attention~\cite{anderson2018bottom, devlin2018bert,lu2016hierarchical} and multi-modal grounding~\cite{antol2015vqa, lu2019vilbert, tan2019lxmert} facilitate the agent to understand the environment. 
Moreover, many reinforcement learning works~\cite{mnih2016asynchronous, jaderberg2017reinforcement, schulman2017proximal} help the agent to obtain a robust navigation policy.
Benefited from these works, previous attempts in the field of Vision-Language Navigation have made great progress in improving the ability to perceive the vision and language inputs~\cite{gupta2017cognitive, Speaker-Follower, wang2018look, RCM+SIL}, and learning a robust navigation policy~\cite{AuxRN, PREVALENT, RVLN}.
However, the VLN task still contains large bias due to the disparity ratio between small data scale and large navigation space, which impacts the generalization ability of navigation. Although the mostly widely used dataset, Room-to-room dataset~\cite{Seq2Seq}, contains only 22K instruction-path pairs, the actual possible navigation path space increases exponentially along with the path length. Thus, the learned navigation policy can easily overfit to the seen scenes and is hard to generalize to the unseen scenes. 

Previous works have proposed various data augmentation methods in an attempt to reduce data bias. 
Fried \emph{et al.} propose a speaker-follower framework~\cite{Speaker-Follower} to generate more data pairs in order to reduce the data bias of the data samples. 
Tan~\emph{et al.}~\cite{EnvDrop} propose an environmental dropout method to augment the vision features in environments; thereby reducing the vision bias inside a house scene. 
However, these methods focus on intra-scene data augmentation and fail to explicitly reduce the data bias across different house scenes.


Accordingly, in this paper, we propose to reduce the domain gap across different house scenes by means of scene-wise data augmentation.
If an agent sees different house scenes during a navigation process, it will be less likely to overfit to a part of the scene textures or room structures. Inspired by this motivation, we propose our method, named Random Environmental Mixup (REM), to improve the generalization ability of a navigation agent. 
REM breaks up two scenes and the corresponding paths, followed by recombining them to obtain a cross-connected scene between the two scenes. 
The REM method provides more generalized data, which helps reduce the generalization error, so that the agent’s navigation ability in the seen and unseen scenes can be improved.

The REM method comprises three steps. 
First, REM selects the key vertexes in the room connection graph according to the betweenness centrality~\cite{brandes2001faster}. 
Second, REM splits the scenes by the key vertexes and cross-connect them to generate new augmented scenes. We propose an orientation alignment approach to solve the feature mismatch problem. 
Third, REM splits trajectories and instructions into sub-trajectories and sub-instructions by their context, then cross-connects them to generate augmented training data. 
An overview of the REM method is presented in Fig.~\ref{fig:introduction}. 

The experimental results on benchmark datasets demonstrate that REM can significantly reduce the performance gap between seen and unseen environments, which dramatically improves the overall navigation performance. 
Our ablation study shows that the proposed augmentation method outperforms other augmentation methods at the same augmentation data scales. 
Our final model obtains 59.1\% in Success weighted by Path Length~(SPL)~\cite{anderson2018on}, which is 2.4\% higher than the previous state-of-the-art result; accordingly, our method becomes the new state-of-the-art method on the standard VLN benchmark. 

\section{Related Work}
\noindent\textbf{Embodied Navigation Environments} are attracting rising attention in artificial intelligence. House3D~\cite{wu2018building} is a manually created large-scale environment. AI2-THOR~\cite{kolve2017ai2} is an interactable indoor environment. Agents can interact with certain interactable objects, such as opening a drawer or picking up a statue. Recent works have tended to focus on simulated environments based on real imagery. The Active Vision dataset~\cite{ammirato2017a} consists of dense scans of 16 different houses. Moreover, Matterport3D~\cite{Seq2Seq}, Gibson~\cite{xia2018gibson} and Habitat~\cite{savva2019habitat} propose high-resolution photo-realistic panoramic view to simulate more realistic environment.

\noindent\textbf{Vision-Language Navigation} has attracted widespread attention, since it is both widely applicable and a challenging task. 
Anderson \emph{et al.}~\cite{Seq2Seq} propose the Room-to-Room (R2R) dataset, which is the first Vision-Language Navigation (VLN) benchmark to combine real imagery~\cite{chang2017matterport3d} and natural language navigation instructions. 
In addition, the TOUCHDOWN dataset~\cite{chen2019touchdown} with natural language instructions is proposed for street navigation. 
To address the VLN task, Fried \emph{et al.} propose a speaker-follower framework~\cite{Speaker-Follower} for data augmentation and reasoning in a supervised learning context, along with a concept named ``panoramic action space'' that is proposed to facilitate optimization. Wang \emph{et al.}~\cite{RCM+SIL} demonstrate the benefit of combining imitation learning~\cite{bojarski2016end, ho2016generative} and reinforcement learning~\cite{mnih2016asynchronous, schulman2017proximal}. Other methods~\cite{wang2018look, ma2019self, ma2019the, EnvDrop, FAST-Short, huang2019transferable} have been proposed to solve the VLN tasks from various perspectives. 
Due to the success of BERT~\cite{devlin2018bert}, researchers have extended it to learn vision-language representations in VLN. PRESS~\cite{PRESS} applies the pre-trained BERT to process instructions. PREVALENT~\cite{PREVALENT} pre-trains an encoder with image-text-action triplets to align the language and visual states, while VLN-BERT~\cite{majumdar2020improving} fine-tunes ViLBERT~\cite{lu2019vilbert} with trajectory-instruction pairs. Hong \emph{et al.}~\cite{RVLN} implements a recurrent function to leverage the history-dependent state representations based on previous models.

\noindent\textbf{Data Augmentation} is widely adopted in diverse deep learning methods. 
Early data augmentation methods in the field of computer vision were manually designed; these include distortions, scaling, translation, rotation and color shifting~\cite{cirean2012multi, wan2013regularization, simard2003best, sato2015apac}. The traditional approach to text augmentation tends to involve a primary focus on word-level cases~\cite{zhang2015character, kobayashi2018contextual, wei2019eda}. 
Some works have achieved success in using GAN to generate augmentation data~\cite{bowles2018gan, sandfort2019data}. 
Zhang~\emph{et al.}~\cite{zhang2017mixup} propose mixup, which is a linear interpolation augmentation method used to regularize the training of neural networks. 
Data augmentation has also been investigated in RL, including domain randomization~\cite{tobin2017domain, pinto2018asymmetric, sadeghi2017cad2rl}, cutout~\cite{cobbe2019quantifying} and random convolution~\cite{lee2020network}. 
In the vision-language navigation context, Fried~\emph{et al.} propose to use a generation method~\cite{Speaker-Follower} to generate data pairs, while Tan~\emph{et al.}~\cite{EnvDrop} propose an environmental dropout method to augment the vision features in various environments. In a departure from these augmentation methods, our REM method cross-connects the scenes and the instruction-trajectory pairs, thereby improving the model's generalization ability among different scenes.

\section{Preliminaries}

\subsection{Vision-language Navigation}
\label{sec:ProblemSetup}
Given a series of triples (environment $E$, path $P$, instruction $I$), the VLN task requires the agent to understand the instruction in order to find a matching path in the corresponding environment. The environment $E$ contains a large number of seen and unseen scenes, the path $P=\{p_0,\dots,p_n\}$ is composed of a list of viewpoints with a length of $n$; morever, instruction $I=\{w_0,\dots,w_m\} $ consists of $m$ words, and a certain correspondence exists between path $P$ and instruction $I$. At time step $t$, the agent observes panoramic views $O_t=\{o_{t,i}\}_{i=1}^{36}$ and navigable viewpoints (at most $k$).
The picture $O_t$ is divided into 12 views in the horizontal direction and 3 views in the vertical direction, for a total of 36 views. 
At the $t$-th step, the agent predicts an action $a \sim \pi_\theta(I,O_t)$, where $\pi_\theta$ is the policy function defined by the parameter $\theta$. 
The actions include `turn left', `turn right' and `move forward' as defined in the Matterpot environment~\cite{Seq2Seq}.
In the Matterport dataset~\cite{chang2017matterport3d}, each scene is discretisized by a navigation graph consists of viewpoints. We model each scene as a graph $G=(V,E)$, where the vertexes set $V$ is a set of scene viewpoints, while $E$ is the connection between viewpoints.

\subsection{Reduce Generalization Error in VLN}
\label{sec:Reduce Generalization Error}
The learning problem can be formulated as the search of the function $f\in \mathcal{F}$, which minimizes the expectation of a given loss $\ell(f(x),y)$. However, the distribution $\mathcal{P}$ of sample $(x,y)$ is generally unknown. We can often obtain a set $\mathcal{T}\sim \mathcal{P}$ and use it as a training set. The approximate function $\hat{f}$ can then be implemented by Empirical Risk Minimization (ERM)~\cite{vapnik1998statistical}.
However, a gap still exists between $\hat{f}$ and $f$. This error describes the generalization ability of $\hat{f}$. The generalization error can be expressed as follows: 
\begin{align}
   \begin{split}
   R_{ge}(\hat{f})&=R(f)-R(\hat{f})\\&=\int(\ell(f(x),y)-\ell(\hat{f}(x),y))d\mathcal{P}(x,y).
   \end{split}
   \label{eq:GE}
\end{align}
In order to enhance the generalization ability of $\hat{f}$, it is necessary to reduce $R_{ge}$.  
According to the Vicinal Risk Minimization (VRM)~\cite{chapelle2001vicinal}:
\begin{align}
   R_v(f)=\frac{1}{m}\sum_{i=1}^n\ell(f(\tilde{x}),\tilde{y}),
   \label{eq:VRM}
\end{align}
where $(\tilde{x},\tilde{y})\in(\mathcal{T}\cup\tilde{\mathcal{T}})$, $\tilde{\mathcal{T}}\sim \mathcal{P}$, $\tilde{\mathcal{T}}\nsubseteq \mathcal{T}$. This means that more samples are needed to lower $R_{ge}$. When the number of samples is certain, the farther the distance $d((\tilde{x},\tilde{y}),\mathcal{T})$ from the sample $(\tilde{x},\tilde{y})$ to the training set $\mathcal{T}$, the better the generalization ability. 

\begin{figure}[t]
   \begin{center}
   \includegraphics[width=0.95\linewidth]{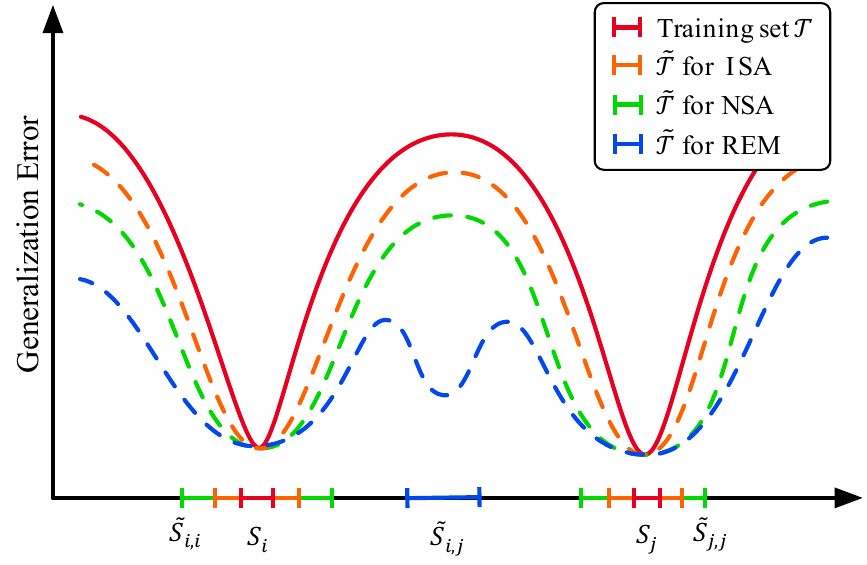}
   \end{center}
   \vspace{-1.2 em}
      \caption{The generalization error of the original training set and different augmentation training sets. With 
      Original (\textbf{Red}) $\rightarrow$ ISA (\textbf{Orange}) $\rightarrow$ NSA (\textbf{Green}) $\rightarrow$ REM (\textbf{Blue}), the distance between $\mathcal{T}$ and $\tilde{\mathcal{T}}$ is getting farther and farther, and the generalization error decreases accordingly.
      } 
      \vspace{-1.2 em}
   \label{fig:GE}
\end{figure}

In the VLN task, the training set consists of $N$ scenes:
\begin{align}
   \mathcal{T}=(S_1\cup S_2 \cup \dots \cup S_N).
\end{align}
We define a data augmentation function $\mathrm{aug}(S_i,S_j)$. 
The generated augmentation data follows the distribution $\mathcal{P}$:
\begin{align}
  \begin{split}
      &\tilde{S}_{i,i}=\mathrm{aug}(S_i, S_i)\sim \mathcal{P},\\
      &\tilde{S}_{i,j}=\mathrm{aug}(S_i,S_j)\sim \mathcal{P},
  \end{split}
  \label{eq:VLN}
\end{align}
where $\tilde{S}_{i,i}$ is the intra-scene augmentation data set, and $\tilde{S}_{i,j}$ is the inter-scene augmentation data set. According to Eqs.~\ref{eq:GE}~and~\ref{eq:VRM}, we have the following assumption:
compared with $\tilde{S}_{i,i}$, the distance from $\tilde{S}_{i,j}$ to $S_i$ is farther, denoted as $d(\tilde{S}_{i,i},S_i)<d(\tilde{S}_{i,j},S_i)$. 
Therefore, the model learned on the inter-scene augmentation data has a smaller generalization error than which learned on the intra-scene augmentation data. 

Previous methods have proposed two kinds of data augmentation methods in VLN: 
the intra-scene augmentation (ISA) method, as in~\cite{Speaker-Follower}, only constructs new paths and instructions in the scene; the near-scene augmentation (NSA) method, as in~\cite{EnvDrop},  breaks through the limitations of the scene to a certain extent by adding Gaussian noise to the scene, but only expands the scene to a small neighborhood. For our part, we propose a inter-scene data augmentation method: Random Environmental Mixup (REM). REM method mixes up two scenes constructs a cross-connected scene between the two scenes. In contrast to the other methods, it exceeds the limitation of the scene itself and constructs augmentation data under a broader data distribution.

Fig.~\ref{fig:GE} illustrates the difference between three methods. The inter-scene method provides more generalized data; this helps to reduce the generalization error, meaning that the agent's navigation ability in the seen scene and the unseen scene can be improved. Subsequent experiments have verified this assumption.


\begin{center}
   \begin{algorithm}[t]
      \label{alg:keynode}
      \caption{Selecting key vertexes}
      \LinesNumbered
      \KwIn{Scene graph $G$; Paths list $P=\{p_1,...,p_{|P|}\}$}
      \KwOut{Key vertexes $v_s^{key},v_t^{key}$}
     Get vertexes set $V$ of $G$\;
     Get edges set $E$ of $G$\;
     $\tilde{V} = \{v \ | \ $top 10 $v$ in $V$ ordered by $VC_B(v)\}$\;
     $\tilde{E} = \{e \ | \ $top 10 $e$ in $E$ ordered by $EC_B(e)\}$\;
      $m=0$\;
      // Select the vertex passed by the most paths.  \\
      \For{$e$ in $E$}{
         Get vertexes $v_s, v_t$ of the edge $e$\;
         $n_e=\sum_{i=1}^{|P|} ({1}_{\{e \in p_i\}}+{0}_{\{e \notin p_i\}})$\;
         \If{$v_s, v_t \in V$}{
            \If{$m < n_e$}{
               $m = n_e$\;
               $v_s^{key} = v_s$\;
               $v_t^{key} = v_t$\;
            }
         }
      }
      \Return{$v_s^{key},v_t^{key}$}
     \end{algorithm}
     \vspace{-1.6 em}
     \end{center}

\section{Random Environmental Mixup}
We propose an inter-scene data augmentation method to construct new environments, paths, and instructions with the aid of training sets. In the training set of the VLN task, there are a large number of different scenes. We randomly select two scenes from the set of training scenes and mix them up to generate cross-connected scenes. 
Adopting this approach enables us to construct the corresponding paths and instructions.
When mixing of scenes, we have the following problems: 
1) how to choose key vertexes in the scene for mixup?
2) how to mix up two scenes to obtain cross-connected scenes?
3) how to construct new paths and instructions in cross-connected scenes?
Solutions to these problems are presented below construct a large number of cross-connected scenes, that are unseen relative to the original training set. 

\subsection{Select Key Vertexes}
\label{sec:SelectingKeyNodes}
\begin{figure}[t]
   \begin{center}
   \includegraphics[width=0.95\linewidth]{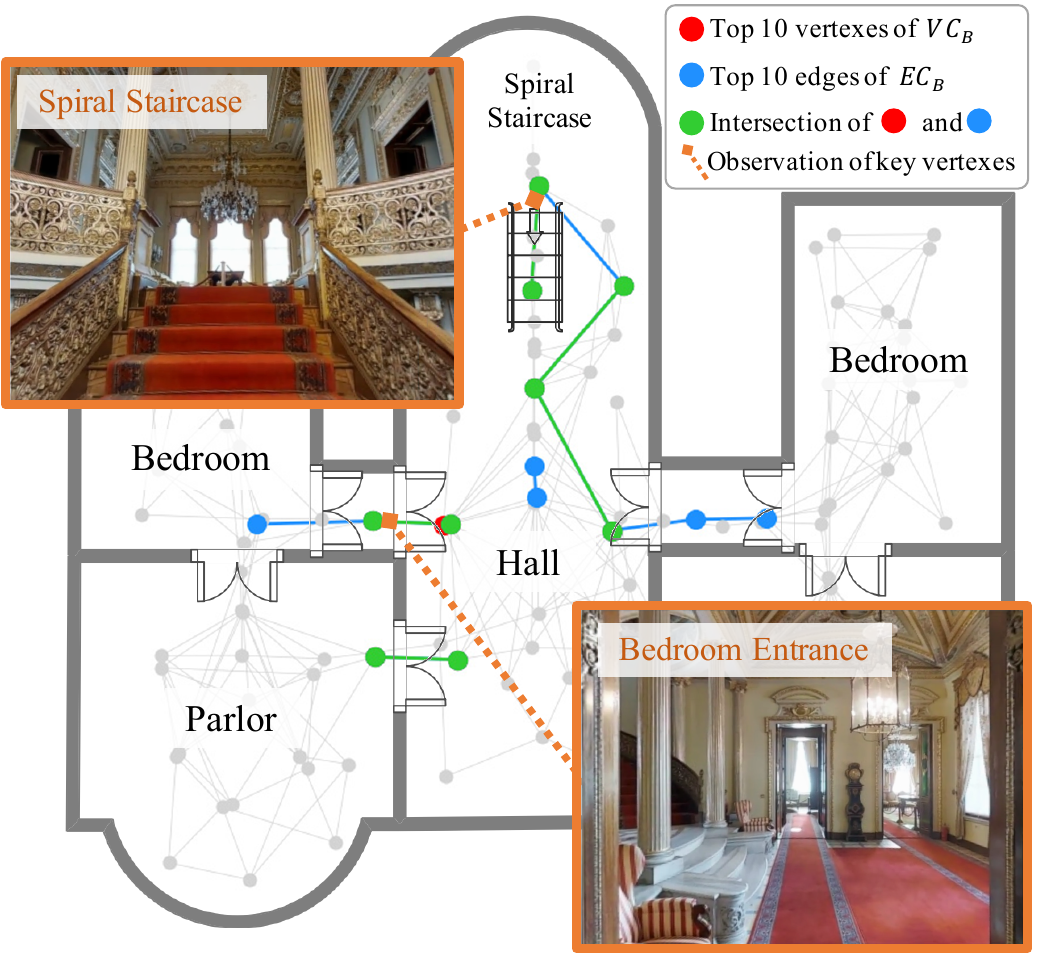}
   \end{center}
   \vspace{-0.4 em}
      \caption{Selecting key vertexes through betweenness centrality.
      The green edges are often the entrances and exits of rooms or corridors, we choose the \textbf{two vertexes} of the green edge that contains the most paths as the key vertexes.} 
      \vspace{-1 em}
   \label{fig:keynode}
\end{figure}
Key vertexes are crucial to mixing between scenes. Their characteristics can be summarized as follows:
1) the entrance or corridor that connects two rooms;
2) the vertex has many paths through it.  
In order to match the above characteristics, key vertexes can be selected with reference to the betweenness centrality~\cite{brandes2001faster} of the graph:  
\begin{align}
\begin{split}
   VC_B(v)&=\sum_{s\neq v \neq t\in V}\frac{\sigma_{st}(v)}{\sigma_{st}},\\
   EC_B(e)&=\sum_{s\neq t\in V;e\in E}\frac{\sigma_{st}(e)}{\sigma_{st}},
\end{split}
\end{align}
where $VC_B(v)$ is the betweenness centrality of the vertex $v$, $EC_B(e)$ is the betweenness centrality of the edge $e$; $\sigma_{st}(v)$ is the number of shortest paths from $s$ to $t$, passes through the vertex $v$; $\sigma_{st}(e)$ is the number of shortest paths from $s$ to $t$ through edge $e$; $\sigma_{st}$ is the number of all shortest paths from $s$ to $t$. Betweenness centrality describes the importance of vertex by the number of shortest paths passing through vertexes or edges. 
Once the vertex is removed from the graph, the points on both sides will be disconnected. 
\begin{figure*}[t]
     \begin{center}
      \vspace{-0.8 em}
     \includegraphics[width=1\linewidth]{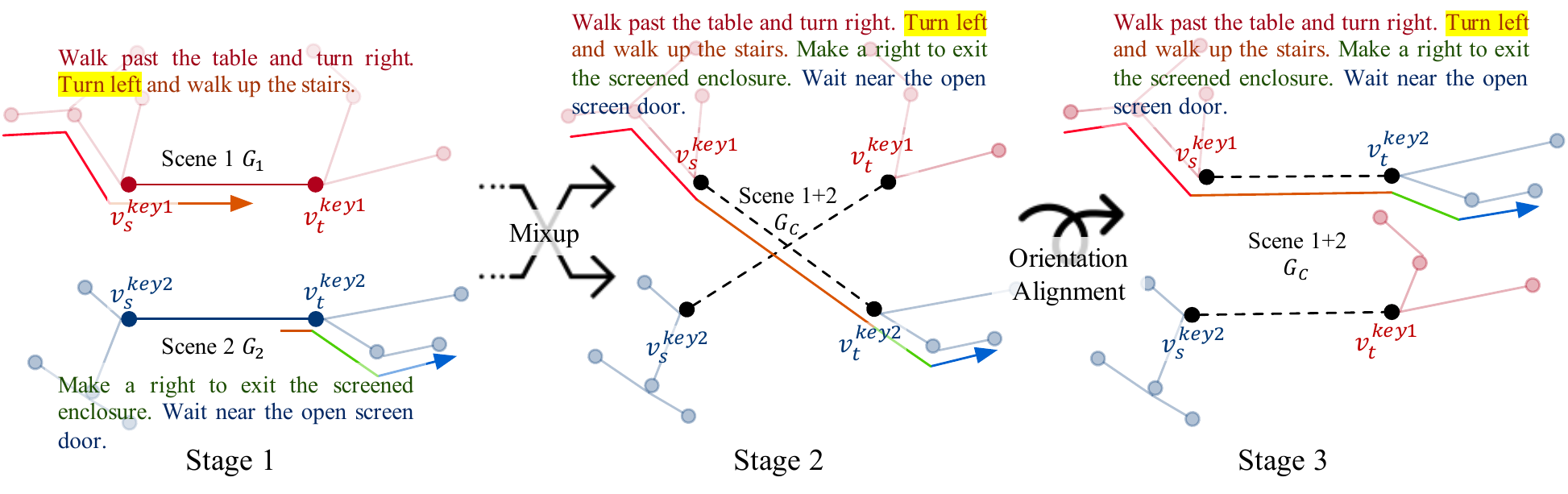}
     \end{center}
     \vspace{-1.2 em}
        \caption{Three stages of mixup scenes. \textbf{Stage 1:} select key vertexes $(v_s^{key1},v_t^{key1})$ and $(v_s^{key2},v_t^{key2})$ for scene 1 and scene 2. \textbf{Stage 2:} mixup scene 1 and scene 2, relink $(v_s^{key1},v_t^{key2})$ and $(v_s^{key2},v_t^{key1})$. \textbf{Stage 3:} fix the position of the vertexes, align the orientation of $(v_s^{key1},v_t^{key2})$ and $(v_s^{key2},v_t^{key1})$. The instructions are fine-grained, and the sub-paths of different colors are matched with the sub-instructions of the corresponding colors. As the scene is mixed up, paths and instructions are also broken up and reconstructed. The constructed scenes, paths and instructions are combined into triples, which become augmentation data for VLN tasks.} 
     \label{fig:crossscene}
     \vspace{-0.8 em}
  \end{figure*}

\begin{figure}[t] 
   \begin{center}
   \includegraphics[width=1\linewidth]{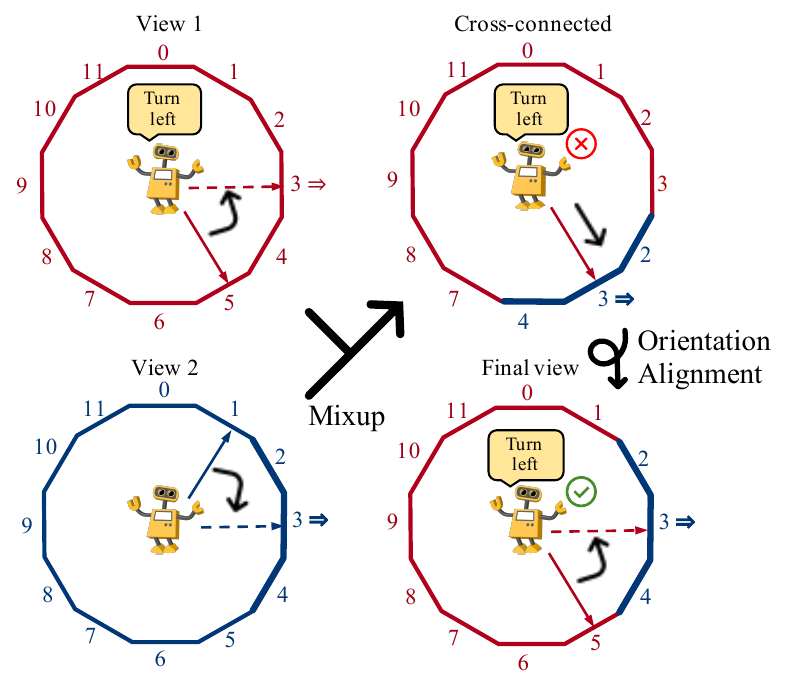}
   \end{center} 
   \vspace{-1.2 em}
      \caption{The process of mixing up viewpoints. \textbf{View~1} is the visual observation of $v_s^{key1}$ in Scene~1, while \textbf{View~2} is the visual observation of $v_s^{key2}$ in Scene~2. The solid arrow is the current direction of the agent. The dotted arrow is the direction of the agent after taking the action. `$\Rightarrow$' represents the direction to the next viewpoint. `turn left' is the instruction received by the agent. `$\Lsh$,$\rightarrow$' are the `turn left' and `forward' actions taken by the agent in order to arrive at the next viewpoint.} 
   \label{fig:viewpoint}
   \vspace{-0.8 em}
\end{figure}

As shown in Fig.~\ref{fig:keynode}, we select the top 10 vertexes and edges of betweenness centrality to obtain the corresponding sets $V^{VC_B}$ and $E^{EC_B}$; subsequently, by excluding edges in $E^{EC_B}$ whose vertexes are not in $V^{VC_B}$, we obtain the final key subgraph $G^{C_B}$. In order to ensure that more paths are subsequently generated, we select the edge $e^{key}$ that contains the most supervised paths from $G^{C_B}$, along with its corresponding vertexes $v_s^{key},v_t^{key}$. 
We observe from Fig.~\ref{fig:keynode} that the entrances and exits of rooms or corridors often have the highest betweenness. 
The process of selecting key vertexes is summarized in Algo.~\ref{alg:keynode}. 

\subsection{Construct Augmented Triplets}
\label{sec:ConstructCrossScenes}
\noindent \textbf{Construct Cross-Connected Scenes\ }
We randomly select two scenes (Scene1 $G_1$ and Scene2 $G_2$) in the training set. We construct the cross-connected scene $G_C$ for $G_1$ and $G_2$ in three stages (Fig.~\ref{fig:crossscene}). In stage 1: according to Algo.\ref{alg:keynode}, we obtain the key vertexes $(v_s^{key1},v_t^{key1})$ for $G_1$ and $(v_s^{key2},v_t^{key2})$ for $G_2$. In stage 2: we mix up $G_1$,$G_2$ into graph $G_C$, disconnect the two key edges $e^{key1}$,$e^{key2}$, and link $(v_s^{key1},v_t^{key2})$,$(v_s^{key2},v_t^{key1})$. In this way, we obtain a cross-connected scene $G_C$. In stage 3: we align the orientation of $G_C$; ensure the matching of the cross path and the instruction by adjusting the vertex position in $G_C$. 

\noindent \textbf{Construct Cross Viewpoints\ }
$G_C$ is a graph containing the connection relationship information without visual observation. 
Therefore, we build a cross viewpoint on the basis of $G_C$ to obtain a new cross-connected environment. 
The process of building a new cross-connected environment is illustrated in Fig.~\ref{fig:viewpoint}. 
Taking the $v_s^{key1}$ in Scene 1+2 as an example, as described in Sec.~\ref{sec:ProblemSetup}, each viewpoint panoramic view is divided into 12 views in the horizontal direction (indicated by the numbers 0–11). By mixing the views of View 1 and View 2, we can obtain a panoramic view of View 1+2.
More specifically, the view is based on the direction of the next viewpoint.
We replace three views around the original angle of View 2 with View 1 to get 
Cross-connected view({\color{red}red 0–3 7–11} from View 1, {\color{blue}blue 2-4} from View 2). The hyperparameter settings for replacing 3 views will be discussed in the experimental part.
 
\begin{table*}[t]
   \begin{center}
   \resizebox{0.96\textwidth}{!}{
   \setlength{\tabcolsep}{0.6em}
   \begin{tabular}{l|c c c c|c c c c|c c c c}
   \hline\hline
   \multirow{2}*{Method} & \multicolumn{4}{|c|}{R2R Validation Seen} & \multicolumn{4}{|c|}{R2R Validation Unseen} & \multicolumn{4}{|c}{R2R Test Unseen} \\
   \cline{2-13}
   ~ & TL & NE$\downarrow$ & SR$\uparrow$ & SPL$\uparrow$ & TL & NE$\downarrow$ & SR$\uparrow$ & SPL$\uparrow$ & TL & NE$\downarrow$ & SR$\uparrow$ & SPL$\uparrow$\\
   \hline\hline
   Random                     &  9.58 & 9.45 & 16 & -  &  9.77 & 9.23 & 16 & -  &  9.89 & 9.79 & 13 & 12 \\
   Human                      & -     & -    & -  & -  & -     & -    & -  & -  & 11.85 & 1.61 & 86 & 76 \\
   \hline
   Seq2Seq-SF~\cite{Seq2Seq}                 & 11.33 & 6.01 & 39 & -  &  8.39 & 7.81 & 22 & -  &  8.13 & 7.85 & 20 & 18 \\
   Speaker-Follower~\cite{Speaker-Follower}           & -     & 3.36 & 66 & -  & -     & 6.62 & 35 & -  & 14.82 & 6.62 & 35 & 28 \\
   SMNA~\cite{SMNA}                      & -     & 3.22 & 67 & 58 & -     & 5.52 & 45 & 32 & 18.04 & 5.67 & 48 & 35 \\
   RCM+SIL~\cite{RCM+SIL}            & 10.65 & 3.53 & 67 & -  & 11.46 & 6.09 & 43 & -  & 11.97 & 6.12 & 43 & 38 \\
   PRESS~\cite{PRESS}                     & 10.57 & 4.39 & 58 & 55 & 10.36 & 5.28 & 49 & 45 & 10.77 & 5.49 & 49 & 45 \\
   FAST-Short~\cite{FAST-Short}                 & -     & -    & -  & -  & 21.17 & 4.97 & 56 & 43 & 22.08 & 5.14 & 54 & 41 \\
   EnvDrop~\cite{EnvDrop}                    & 11.00 & 3.99 & 62 & 59 & 10.70 & 5.22 & 52 & 48 & 11.66 & 5.23 & 51 & 47 \\
   AuxRN~\cite{AuxRN}                      & -     & 3.33 & 70 & 67 & -     & 5.28 & 55 & 50 & -     & 5.15 & 55 & 51 \\
   PREVALENT~\cite{PREVALENT}                  & 10.32 & 3.67 & 69 & 65 & 10.19 & 4.71 & 58 & 53 & 10.51 & 5.30 & 54 & 51 \\
   RelGraph~\cite{RelGraph}                   & 10.13 & 3.47 & 67 & 65 &  9.99 & 4.73 & 57 & 53 & 10.29 & 4.75 & 55 & 52 \\
   VLN$\circlearrowright$Bert~\cite{RVLN} & 11.13 & \textbf{2.90} & \textbf{72} & \textbf{68} & 12.01 & \textbf{3.93} & \textbf{63} & \textbf{57} & 12.35 & \textbf{4.09} & \textbf{63} & \textbf{57} \\
   \hline
   IL+RL*~\cite{EnvDrop}                  & 10.25 & 4.91 & 53.8 & 50.7 &  9.38 & 5.89 & 46.2 & 42.5 & 9.58 & 5.88 & 46.4  & 43.3 \\
   \textbf{IL+RL+REM}                  & 10.18 & 4.61 & 58.2 & 55.3 &  9.40 & 5.59 & 48.6 & 44.8 & 9.81 & 5.67 & 48.7  & 45.1 \\
   EnvDrop*~\cite{EnvDrop}                & 10.46 & 3.78 & 64.4 & 62.0 &  9.50 & 5.52 & 51.1 & 47.3 & 11.32 & 5.84 & 50.5 & 46.5 \\
   \textbf{EnvDrop+REM}                & 11.13 & 3.14 & 70.1 & 66.7 & 14.84 & 4.99 & 53.8 & 48.8 & 10.73 & 5.40 & 54.1 & 50.4 \\
   VLN$\circlearrowright$Bert*~\cite{RVLN}& 12.09 & 2.99 & 70.7 & 65.9 & 12.58 & 4.02 & 61.4 & 55.6 & 11.68 & 4.35 & 61.4 & 56.7 \\
   \textbf{VLN$\circlearrowright$Bert+REM}& 10.88 & \textbf{\color{blue}2.48} & \textbf{\color{blue}75.4} & \textbf{\color{blue}71.8} & 12.44 & \textbf{\color{blue}3.89} & \textbf{\color{blue}63.6} & \textbf{\color{blue}57.9} & 13.11 & \textbf{\color{blue}3.87} & \textbf{\color{blue}65.2} & \textbf{\color{blue}59.1} \\
   \hline\hline
   \end{tabular}}
   \end{center}
   \vspace{-0.2 em}
   \caption{Comparison of agent performance on \textbf{R2R} in single-run setting. * reproduced results in my environment.}
   \vspace{-0.4 em}
   \label{tab:result}
\end{table*}

\begin{table*}[t]
   \begin{center}
   \resizebox{0.96\textwidth}{!}{
   \setlength{\tabcolsep}{0.4em}
   \begin{tabular}{l|c c c c c c|c c c c c c}
   \hline\hline
   \multirow{2}*{Method} & \multicolumn{6}{|c|}{R4R Validation Seen} & \multicolumn{6}{|c}{R4R Validation Unseen}\\
   \cline{2-13}
   ~                                       & NE$\downarrow$   & SR$\uparrow$   & SPL$\uparrow$ & CLS$\uparrow$   & nDTW$\uparrow$  & SDTW$\uparrow$ & NE$\downarrow$   &  SR$\uparrow$  & SPL$\uparrow$ & CLS$\uparrow$  & nDTW$\uparrow$ &  SDTW$\uparrow$ \\
   \hline\hline
   Speaker-Follower~\cite{jain2019stay}    & 5.35 & 51.9 & 37.3 & 46.4 & - & -  & 8.47 & 23.8 & 12.2 & 29.6 & - & -     \\
   RCM~\cite{jain2019stay}                 & 5.37 & 52.6 & 30.6 & 55.3 & - & -  & 8.08 & 26.1 & 7.7 & 34.6 & - & -  \\
   PTA~\cite{landi2019perceive}            & \textbf{4.53} & \textbf{58.0} & \textbf{39.0} & \textbf{60.0} & \textbf{58.0} & \textbf{41.0} & 8.25 & 24.0 & 10.0 & 37.0 & 32.0 & 10.0  \\
   EGP~\cite{deng2020evolving}             & -     & -    & -    & -    & -     & -     & \textbf{8.00} & \textbf{30.2} & - & 44.4 & \textbf{37.4} & \textbf{17.5} \\
   BabyWalk~\cite{zhu2020babywalk}             & -     & -    & -    & -    & -     & -     & 8.2  & 27.3 & \textbf{14.7} & \textbf{49.4} & 39.6 & 17.3\\
   \hline
   IL+RL*~\cite{EnvDrop}                   & 5.94 & 35.3 & 32.5 & 37.1 & 38.7 & 26.5 & 8.88 & 31.9 & 18.7 & 32.3 & 31.7 & 12.2 \\
   \textbf{IL+RL+REM}                      & 6.72 & 39.9 & 36.5 & 42.4 & 47.3 & 31.2 & 8.83 & 33.1 & 20.1 & 38.6 & 37.6 & 15.7 \\
   EnvDrop*~\cite{EnvDrop}                 & 5.94 & 42.7 & 39.5 & 40.2 & 41.8 & 29.6 & 9.18 & 34.7 & 21.0 & 37.3 & 34.7 & 12.1 \\
   \textbf{EnvDrop+REM}                    & 5.83 & 46.3 & 43.5 & 45.1 & 49.7 & 33.4 & 8.21 & 37.9 & 25.0 & 42.3 & 39.7 & 18.5 \\
   VLN$\circlearrowright$Bert*~\cite{RVLN} & 4.84 & 55.7 & 46.0 & 47.8 & 55.8 & 37.9 & 6.48 & 42.5 & 32.4 & 41.4 & 41.8 & 20.9 \\
   \textbf{VLN$\circlearrowright$Bert+REM} & {\color{blue}\textbf{3.77}} & {\color{blue}\textbf{66.8}} & {\color{blue}\textbf{57.4}} & 56.8 & {\color{blue}\textbf{61.5}} & {\color{blue}\textbf{41.5}} & {\color{blue}\textbf{6.21}} & {\color{blue}\textbf{46.0}} & {\color{blue}\textbf{38.1}} & 44.9 & {\color{blue}\textbf{46.3}} & {\color{blue}\textbf{22.7}} \\
   \hline\hline 
   \end{tabular}}
   \end{center}
   \vspace{-0.2 em}
   \caption{Comparison of agent performance on \textbf{R4R} in single-run setting. * reproduced results in my environment.}
   \vspace{-1 em}
   \label{tab:r4rresult}
\end{table*}

\noindent \textbf{Construct Cross Paths and Instructions\ }
Cross-connecting the instructions and paths requires the instructions and paths to be fine-grained. 
In order to obtain the fine-grained data, we use Fine-Grained R2R~\cite{FGR2R} to split the instructions and paths, as well as to align the sub-instructions and the sub-paths. 
As shown in Fig.~\ref{fig:crossscene} (stage 3), we obtain the path and instructions in the cross-connected scene by combining the sub-paths before and after the key vertex, along with the corresponding sub-instructions.

\noindent \textbf{Orientation Alignment\ }
Following Fig.~\ref{fig:crossscene} (stages 1 and 2), we construct the cross-connected scenes and corresponding cross viewpoints. Simply connecting $v_s^{key1}$ and $v_t^{key2}$ leads to a mismatch of the related orientations of the vertexes in the cross-connected scenes.; it is therefore necessary to align the orientation of the vertexes.
More specifically, after the scenes and views are mixed, the direction of `$\Rightarrow$' changes (Fig.~\ref{fig:viewpoint} from $90^{\circ}$ to $150^{\circ}$). Correspondingly, to enable it to go to the next viewpoint, the agent's action also changes (from '$\Lsh$' to '$\rightarrow$'). However the instruction is still `turn left’. To solve this problem of mismatch between action and instruction, we need to fix the position on the cross-connected scenes. To achieve this, as shown in Fig.~\ref{fig:crossscene} (stage 3), we move the vertexes $v_t^{key1}$, $v_t^{key2}$ and their associated vertexes, exchanging the position of the two vertexes, meaning that that the relative positions of the key vertexes remain unchanged. Through fixing the vertexes' position, the orientation of `$\Rightarrow$' is aligned (see Fig.~\ref{fig:viewpoint}~final~view). The agent's action and instructions accordingly match again.

\subsection{Augmentation in Vision Language Navigation}

At this point, we have constructed augmented triplets for training: (environment, path, instruction). Our method is able to mix up any two scenes into a new cross-connected scene. We can accordingly generate a large number of new scenes and their corresponding paths and instructions. 

For VLN tasks, we need to export cross-connected scenes for training, including the viewpoints, connection relations and vertex positions. The augmented triplets will be merged directly with the original training set, namely $\mathcal{T}_{aug} = \tilde{\mathcal{T}}\cup\mathcal{T}$; we use $\mathcal{T}_{aug}$ in place of $\mathcal{T}$ in training. The observation features in different directions for the cross viewpoint are derived from different scenes. 

\section{Experiment}

\subsection{Dataset and Evaluation Setup}

\noindent \textbf{Dataset and Simulator\ } We evaluate our agent on the Room-to-Room (R2R)~\cite{Seq2Seq} and R4R~\cite{jain2019stay} based on Matterport3D simulator~\cite{chang2017matterport3d}. This is a powerful navigation simulator. 
R4R builds upon R2R and aims to provide an even more challenging setting for embodied navigation agents. In a scene, the agent will jump between pre-defined viewpoints on the connectivity graph of the environment.

\noindent \textbf{Evaluation Metrics\ }  There are already many recognized indicators used to evaluate models in VLN: Trajectory Length (TL), trajectory length in meters; Navigation Error (NE), error from the target point in meters; Success Rate (SR), the proportion of times that the agent successfully arrived within 3 meters of the target; and the success rate weighted by the path length (SPL)~\cite{anderson2018on}. In R4R, CLS~\cite{jain2019stay}, nDTW and SDTW~\cite{ilharco2019general} take into account the agent’s steps and are sensitive to intermediate errors in the navigation path. 

\noindent \textbf{Implementation Details\ } We use EnvDrop~\cite{EnvDrop} and VLN$\circlearrowright$Bert~\cite{RVLN} as the baselines to evaluate our method. In the interests of fairness, we use the same experimental settings as the original method. On the basis of not changing the hyperparameter settings, augmented triplets are added for training. We randomly paired and mixed up the 61 scenes in the training set, finally obtaining 116 cross-connected scenes, 5916 paths and 7022 instructions.

\begin{table}[t]
   \begin{center}
   \begin{tabular}{l|l|cccc}
   \hline\hline
   & Method    & NE$\downarrow$ & OR$\uparrow$ & SR$\uparrow$ & SPL$\uparrow$ \\ \hline\hline
   \multirow{5}{*}{\rotatebox{90}{Val Seen}}    & Baseline  & 4.91 & 62.3 & 53.8 & 50.7 \\
                                                & Before OA      & 4.83 & 63.1 & 54.6 & 53.2 \\
                                                & Before OA + CCV & 4.72 & 64.3 & 56.8 & 54.1 \\
                                                & After OA      & 4.78 & 63.7 & 55.6 & 53.8 \\
                                                & After OA + CCV & \textbf{4.61} & \textbf{65.6} & \textbf{58.2} & \textbf{55.3} \\\hline\hline
   \multirow{5}{*}{\rotatebox{90}{Val Unseen}}  & Baseline  & 5.89 & 54.5 & 46.2 & 42.5 \\
                                                & Before OA      & 5.92 & 53.0 & 46.0 & 42.4 \\
                                                & Before OA + CCV & 5.88 & 54.8 & 46.9 & 42.8 \\
                                                & After OA      & 5.73 & 55.2 & 47.2 & 43.2 \\
                                                & After OA + CCV & \textbf{5.59} & \textbf{56.0} & \textbf{48.6} & \textbf{44.8} \\\hline\hline
   \end{tabular}
   \end{center}
   \caption{Model performance before and after orientation alignment. \textbf{Before OA} means before orientation alignment; \textbf{After OA} means after orientation alignment; \textbf{CCV} means replacement visual observation in construct cross viewpoints.}
   \label{tab:pos}
   \vspace{-0.8 em}
   \end{table} 

\subsection{Results on VLN Standard Benchmark}

In this section, our method is compared with several other representative methods. We apply the proposed REM to three baseline methods and compare them with other methods. 
Tab.~\ref{tab:result} shows the results on R2R. REM achieves excellent performance on the three baseline methods. In the state-of-the-art method, REM can further improve performance. 
Tab.~\ref{tab:r4rresult} shows the results on R4R. Through REM, all three baseline methods have been significantly improved. In addition to the success rate and SPL, REM can also significantly improve CLS and nDTW, which shows that the proposed method can make the agent follow the instructions and make the navigation path more matched.

\begin{table}[t] 
   \begin{center}
   \begin{tabular}{l|l|cccc}
   \hline\hline
   & Method    & NE$\downarrow$ & OR$\uparrow$ & SR$\uparrow$ & SPL$\uparrow$ \\ \hline\hline
   \multirow{5}{*}{\rotatebox{90}{Val Seen}}    & 0 View   & 4.78 & 63.7 & 55.6 & 53.8 \\
                                                & 1 View   & 4.70 & 64.5 & 56.8 & 54.6 \\
                                                & 2 Views  & 4.64 & 65.1 & 57.2 & 54.9 \\
                                                & 3 Views  & \textbf{4.61} & \textbf{65.6} & \textbf{58.2} & \textbf{55.3} \\
                                                & 4 Views  & 4.67 & 64.0 & 57.6 & 55.0 \\
                                                \hline\hline
   \multirow{5}{*}{\rotatebox{90}{Val Unseen}}  & 0 View   & 5.73 & 55.2 & 47.2 & 43.2 \\
                                                & 1 View   & 5.68 & 55.9 & 47.5 & 43.4 \\
                                                & 2 Views  & 5.63 & \textbf{56.2} & 48.1 & 44.1 \\
                                                & 3 Views  & \textbf{5.59} & 56.0 & \textbf{48.6} & \textbf{44.8} \\
                                                & 4 Views  & 5.66 & 55.3 & 47.9 & 44.3 \\
                                                \hline\hline
   \end{tabular}
   \end{center}
   \caption{The impact of replacing the number of different views on the model performance. "0 View" means that visual observation is not replaced}
   \label{tab:viewpoint}
   \vspace{-0.8 em}
   \end{table}

   \begin{figure*}[]
      \vspace{-0.2 em}
      \begin{center}
      \includegraphics[width=1\linewidth]{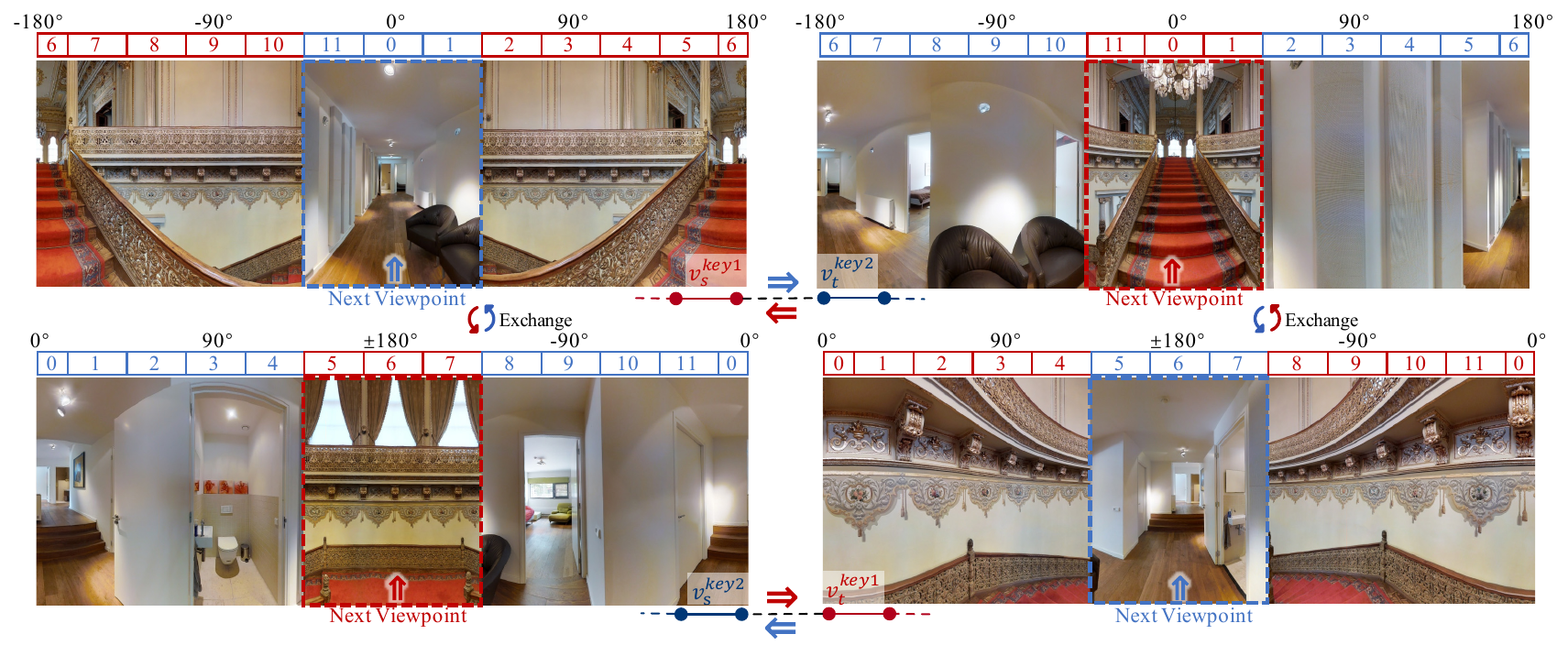}
      \end{center}
      \vspace{-1.6 em}
         \caption{Schematic diagram of cross viewpoints. After the key viewpoints of the two scenes are mixed; the upper and lower two viewpoints' divided views exchange each other; the two viewpoints on the left and right are connected to each other, and the agent can go to each other through `$\Rightarrow$'.} 
      \label{fig:viz}
      \vspace{-1.4 em}
   \end{figure*}
   
   \begin{figure}[]
      \begin{center}
      \includegraphics[width=1\linewidth]{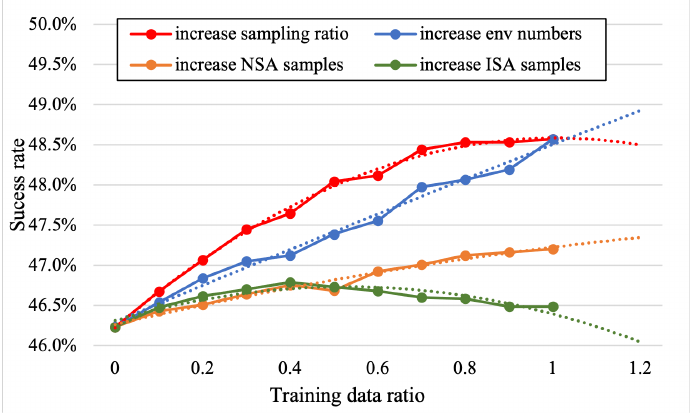}
      \end{center}
      \vspace{-1.1 em}
         \caption{Success rates of agents trained with different amounts of data. The same data ratio in the figure indicates that the same amount of data is used.
        The \textbf{blue} line indicates that the results are increased by gradually adding new environments to the supervised training method. The \textbf{red} line only gradually increases the amount of data and randomly selects data from all training environments.
        } 
      \label{fig:ablation}
      \vspace{-1 em}
   \end{figure}

\subsection{Method Analysis}
\noindent \textbf{Orientation Alignment\ } 
In Sec.~\ref{sec:ConstructCrossScenes}, we propose the orientation alignment operation. 
Tab.~\ref{tab:pos} shows the performance difference between no orientation alignment (Before OA) and orientation alignment (After OA). Orientation alignment increases the success rate of the baseline by 1\%. If the orientation is not aligned, the performance of the model will decrease instead. This is because the agent's actions and instructions do not match, and the agent cannot correctly learn the relationship between instructions and action.
In addition, we tested the effect of replacing visual observation (CCV) on the results, and After OA achieved the highest improvement.

\noindent \textbf{Replace Visual Observation\ } 
In the process of constructing the cross viewpoint, we performed the operation of replacing the visual observation in the specified direction. There are a total of 12 view directions in the horizontal direction. We experimented to determine how many views should be replaced to achieve the best results for the model. Tab.\ref{tab:viewpoint} outlines the effect of replacing different numbers of views on model performance. As the table shows, three views is the optimal choice.
Excessive replacement of visual observation information is thus suboptimal. Through experiments, we choose each cross viewpoint in REM to replace views in three horizontal directions.
Fig.~\ref{fig:viz} shows a schematic diagram of cross viewpoints.

\subsection{Ablation Analysis}

In order to compare the impact of the number of mixup environments on REM performance, we limited the amount of training data (a data ratio of 1 means 7022 instructions), and compared four different settings: 1) as the amount of data increases, the environments number available for mixup also increases in the same proportion; 2) mixup is always used for all environments, but the instructions number generated; 3) NSA is used to generate the same instructions number; 4) ISA is used to generate the same instructions number. The success rate can indicate the generalization ability of different methods. The higher the success rate of the method, the stronger its generalization ability.

The results are shown in Fig.~\ref{fig:ablation}. With the increase of sampled data, all methods achieve improvements in model performance to a certain extent.
When the data ratio is 1, the red and blue dots have the same setting, the red dot reaches the peak of performance; this means that when the number of mixed scenes is fixed, continuing to increase the sample data cannot further reduce the generalization error.
For the blue line, there is no performance degradation trend observed when the data ratio is 1, which shows that increasing the number of mixed scenes can continue to reduce the generalization error. 
The difference between the red-blue and orange-green lines indicates that, when the sample number is the same, the inter-scene data augmentation is significantly better than the intra-scene data augmentation. This verifies the assumption presented in Sec.~\ref{sec:Reduce Generalization Error}. 

\vspace{-.8 em}
\section{Conclusion}
\vspace{-.8 em}

In this paper, we analyze the factors that affect generalization ability and put forward the assumption that inter-scene data augmentation can more effectively reduce generalization errors.
We accordingly propose the Random Environmental Mixup (REM) method, which generates cross-connected house scenes as augmented data via mixuping environment. The experimental results on benchmark datasets demonstrate that REM can significantly reduce the performance gap between seen and unseen environments. Moreover, REM dramatically improves the overall navigation performance. Finally, the ablation analysis verifies our assumption pertaining to the reduction of generalization errors. 

\vspace{-.8 em}
\section*{Acknowledgments}
\vspace{-.8 em}

This work is supported in part by China National 973 program (Grant No. 2014CB340301), Guangdong Outstanding Youth Fund (Grant No. 2021B1515020061), Zhejiang Lab’s Open Fund (No. 2020AA3AB14), CSIG Young Fellow Support Fund, and Australian Research Council (ARC) Discovery Early Career Researcher Award (DECRA) under DE190100626 and Zhejiang Lab’s Open Fund (No. 2020AA3AB14).

{\small
\bibliographystyle{ieee_fullname}
\bibliography{vln}
}
 
\end{document}